\def\year{2022}\relax
\documentclass[letterpaper]{article} % DO NOT CHANGE THIS
\pdfoutput=1
\usepackage{pdfpages}
\usepackage{pdfpages}
\usepackage{aaai22}  % DO NOT CHANGE THIS
\usepackage{times}  % DO NOT CHANGE THIS
\usepackage{helvet}  % DO NOT CHANGE THIS
\usepackage{courier}  % DO NOT CHANGE THIS
\usepackage[hyphens]{url}  % DO NOT CHANGE THIS
\usepackage{graphicx} % DO NOT CHANGE THIS
\usepackage{natbib}  % DO NOT CHANGE THIS AND DO NOT ADD ANY OPTIONS TO IT
\usepackage{caption} % DO NOT CHANGE THIS AND DO NOT ADD ANY OPTIONS TO IT
\DeclareCaptionStyle{ruled}{labelfont=normalfont,labelsep=colon,strut=off} % DO NOT CHANGE THIS
\usepackage{algorithm}
\usepackage{algorithmic}
\usepackage{dblfloatfix}
\usepackage{amsmath,amsfonts,amssymb}

\DeclareMathOperator*{\argmin}{argmin}

\usepackage{newfloat}
\usepackage{listings}
\floatstyle{ruled}
\newfloat{listing}{tb}{lst}{}
\floatname{listing}{Listing}
\title{Using Graph-Aware Reinforcement Learning to Identify Winning Strategies in Diplomacy Games (Student Abstract)}
\author{
    Hansin Ahuja\textsuperscript{\rm 1}, Lynnette Hui Xian Ng\textsuperscript{\rm 2}, Kokil Jaidka\textsuperscript{\rm 3}
}
\affiliations{
    \textsuperscript{\rm 1}Indian Institute of Technology Ropar\\
    \textsuperscript{\rm 2}Carnegie Mellon University\\
    \textsuperscript{\rm 3}National University of Singapore\\
    hansinahuja@gmail.com,
    huixiann@andrew.cmu.edu, 
    jaidka@nus.edu.sg
    
}

\begin{document}

\maketitle

\begin{abstract}
This abstract proposes an approach towards goal-oriented modeling of the detection and modeling complex social phenomena in multiparty discourse in an online political strategy game.
We developed a two-tier approach that first encodes sociolinguistic behavior as linguistic features then use reinforcement learning to estimate the advantage afforded to any player. 
In the first tier, sociolinguistic behavior, such as Friendship and Reasoning, that speakers use to influence others are encoded as linguistic features to identify the persuasive strategies applied by each player in simultaneous two-party dialogues. 
In the second tier, a reinforcement learning approach is used to estimate a graph-aware reward function to quantify the advantage afforded to each player based on their standing in this multiparty setup.
We apply this technique to the game Diplomacy, using a dataset comprising of over 15,000 messages exchanged between 78 users. Our graph-aware approach shows robust performance compared to a context-agnostic setup.
\end{abstract}

\section{Introduction}
\noindent In an increasingly connected world, communication for personal, entertainment and professional reasons is increasingly online. However, most research on online communication has focused on personal and professional contexts, while online coordination in game-based contexts is less understood. For example, in multiparty online games, players interact through textual, audio, and visual signals to coordinate game strategies aimed at ultimate victory. This is partly due to the lack of multimodal datasets to examine the dynamic social processes at play. 

% Prior work in this field studied multimodal cues of the perceived veracity of speech \cite{chen2020acoustic}, and social influence-based disingenuity \cite{husmile}. However, optimal strategies that interlocutors may employ to maximize chances of success, in a deliberate effort to best other people, have not been studied yet. Addressing this gap, this paper examines a recent dataset ~\cite{peskov2020takes} collected from interacting players during ongoing games of Diplomacy, a political strategy game, as it was played online.

In this work, we study optimal strategies that interlocutors may employ to maximize chances of success, in a deliberate effort to best other people. This paper examines a recent dataset ~\cite{peskov2020takes} collected from interacting players during ongoing games of Diplomacy, a political strategy game, as it was played online.

\section{Approach}
We adopt a two-tier methodology in predicting the winner of a Diplomacy chat thread. First, we relied on sociolinguistic cues from words and word features to infer persuasive strategies such as Friendship and Reasoning. Next, a reinforcement learning approach was used to construct a graph-aware reward function that considers the in-game dynamic between two players and the interplay of players in a more extensive multiparty setup.

\textbf{Dataset}: The CL-Aff Diplomacy dataset~\cite{jaidka2021} comprises additional labels to the Diplomacy dataset~\cite{peskov2020takes} identifying the rhetorical strategies used by the players in their chat messages.
The dataset included four annotated labels about the rhetorical strategies followed by the players (Friendship, Reasoning, Game Move, and Share Information), which had a pairwise inter-annotator agreement of at least 60\%. %The labels are described in Table~\ref{tab:diplomacyannotation}.
Rather than utterance-level deception, our paper is interested in examining whether signals of influence and persuasion applied in the game's early stages predict ultimate victory. 

% \vspace{-0.15cm}

\subsection{Weakly Supervised Labeling of Player Strategies}
The first step involved identifying the action space for the players in the Diplomacy games in terms of the rhetorical strategies that they follow to influence and persuade each other. 
However, only 60\% of the CL-Aff Diplomacy dataset had high-quality labeled data. Hence, a weakly supervised approach was followed to predict the rhetorical strategies for the entire dataset. The training set comprising the labels from CL-Aff Diplomacy was used to train binary classifiers on the different rhetorical strategies, such as Friendship (F), Reasoning (R), Game Move (GM), and Share Information (SI). The best-performing classifiers (reported in Table~\ref{tab:individualscoresdataset1}) were used to predict labels for the entire dataset.\footnote{More results can be found at https://tinyurl.com/diplomacy-rl/} 

% \vspace{-0.15cm}

\subsection{Score-Based Inverse Reinforcement Learning}
In the second step, we identified the winning player in each conversation as a function of their rhetorical strategies. We formulate this as an inverse RL problem and, more specifically, use score-based inverse reinforcement learning (SBIRL) \cite{el2016score}, which allows us to exploit the experiences of non-expert agents as well. 

Each state \(s\) of the state space \(S\) is encoded using \(\phi: S \rightarrow \mathbb{R}^d \). The encoding is done by picking a set of characteristic features corresponding to that state. Simpler operationalizations rely only on the player's score at any given point. On the other hand, a graph-aware operationalization incorporated the player's importance and influence in the textual communication network for the game through several centrality features. For the reward function, a linear parameterization was adopted: \(r_{\theta}(s) = \theta^\intercal \phi(s)\) where \(\theta\) is the set of parameters. Each thread \(t\) of conversation is represented as: 

% \vspace{-0.21cm}

% \[r_{\theta}(s) = \theta^\intercal \phi(s)\]

% \vspace{-0.07cm}

% \vspace{-0.18cm}

\[t = (s_0, \dots , s_{T}) = (s_j)_{j = 0}^{T}\]

% \vspace{-0.05cm}

We extracted the thread-level tuples \((t_i, f_i)\), where  \(t_i\) is the subsequence of states corresponding to the \(i^{th}\) player (referred to as a subthread) and \(f_i\) is the final score of the \(i^{th}\) player at the end of the thread. 
For any given subthread and reward, the discounted sum of rewards can be written as

% \vspace{-0.3cm}

\[\sum_{t = 0}^{T} \gamma^{t} r_{\theta}(s_t) = \theta^\intercal \mu(h) \ \ \textrm{with} \ \ \mu(h) = \sum_{t = 0}^{T} \gamma^{t} \phi(s_t)\]

% \vspace{-0.1cm}

Where \(\gamma\) is the discounting factor. We regress the final scores \(f_i\) on the mappings \(\mu(h_i)\) and asymptotically minimise the risk based on the \(\ell_2\)-loss. A reward function estimator
\(r_{\theta_n}\) is derived after estimating \(\theta_n\) by solving: 
% \vspace{0.65cm} 
% \vspace{-0.1cm}
\[\theta_n = \argmin_{\theta \in \mathbb{R}^d} \frac{1}{n} \sum_{i = 1}^{n} (f_i - \theta^{\intercal} \mu(h_i))^2\]

% \vspace{-0.3cm}
\section{Preliminary Results}
\subsection{Weakly Supervised Labeling of Player Strategies}
Once the annotation labels were obtained, key player strategies were identified. At the player level, Reasoning (R), Game move (GM), and Share Information (SI) all shared a strong pairwise Pearson correlation (\(r \hspace{-0.14cm} \in \hspace{-0.14cm} (0.45, 0.55)\)), while each were strongly anti-correlated with Friendship (F) (\(r \hspace{-0.04cm} \in \hspace{-0.04cm} (\textrm{-}0.61, \textrm{-}0.45)\)), both with \(p < 0.001\). This suggests that players either acted on a Friendship or a Reasoning strategy and enabled us to label each utterance as an action under assumptions of mutual exclusion. When both labels were present, a voting mechanism with the Game move and Share Information was used to determine the ultimate label.
% \vspace{-0.1cm}
\begin{table}[!ht]

% \vspace*{-\baselineskip}
% \vspace{-0.6cm}

\begin{center}
\fontsize{9}{10}\selectfont
% \resizebox{9pt}{!}{
% \begin{table}[b]
\begin{tabular}{l r r r r}

\hline

% & \textbf{Friendship} & \textbf{Reasoning} & \textbf{Game Move}  & \textbf{Share Info}\\ \hline
& \textbf{F} & \textbf{R} & \textbf{GM}  & \textbf{SI}\\ \hline
Logistic regression & 0.633 & 0.508 & 0.695 & 0.541 \\ 
%KNN & 0.570 & 0.471 & 0.601 & 0.574 & 0.629 & 0.496 \\ 
Gaussian NB & 0.467 & \textbf{0.533} & 0.620 & 0.315 \\ 
%Bernoulli NB & 0.607 & 0.520 & 0.642 & 0.635 & 0.676 & 0.544 \\ 
Adaboost & 0.642 & 0.448 & 0.696  & \textbf{0.623} \\ 
Gradient boosting & 0.641 & 0.445 & \textbf{0.698} & 0.611 \\ 
%Decision Tree & 0.603 & 0.502 & 0.644 & 0.600 & 0.643 & 0.575 \\ 
% Linear SVC & 0.627 & 0.524 & 0.695 & 0.678 & 0.688 & 0.560 \\ 
C-SVC & \textbf{0.673} & 0.526 & 0.692  & 0.570 \\ 
%LDA & 0.632 & 0.444 & 0.696 & 0.650 & 0.701 & 0.591 \\ 
\hline
\end{tabular}
% \end{table}
% }
% \vspace{-0.25cm}

\caption{Macro-F1 scores for feature estimators}
\label{tab:individualscoresdataset1}
\end{center}

\end{table}
% \vspace*{-\baselineskip}
% \vspace{-0.8cm}
\subsection{Efficacy of the Reward Function}
The winner of a chat thread is the player with the higher score at the end of the thread. The accuracy of the reward function is the fraction of times when the average estimated reward for the winner was greater than that of the loser. 
% (reported in Table~\ref{tab:sbirl})

\begin{figure}[!ht]
    \centering
    \includegraphics[scale=0.5]{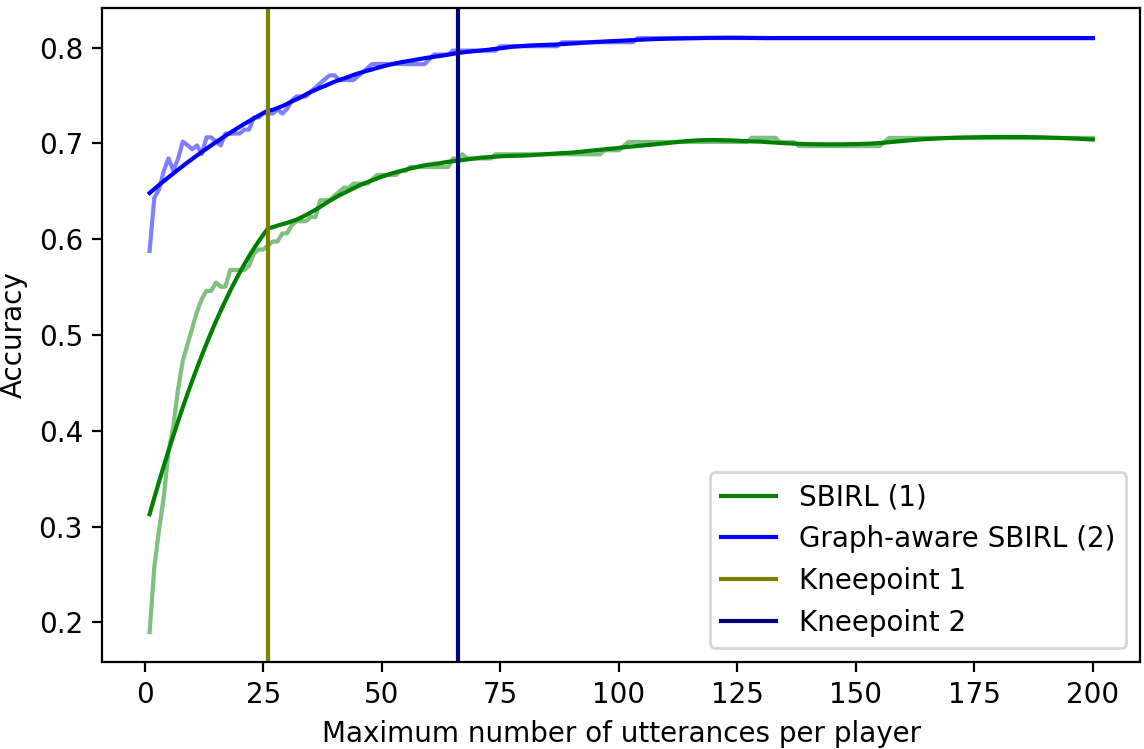} 
    % \vspace{-0.42cm}
    \caption{Accuracies with number of utterances restricted}
  %  \vspace*{-\baselineskip}
    \label{fig:ablation}
\end{figure}

% We have presented the results based on two experimental approaches. 
First, we encoded only the difference between the game scores of the two players in our state feature vector. Next, given that each player is engaged in multiple conversations at the same time, we introduced multiple graph centrality features, such as authority score and eigen centrality, into the state representation\footnote{Calculated using igraph, https://igraph.org/r/}. The results are reported in Table~\ref{tab:sbirl}. We note that our graph-aware approach outperforms the context-agnostic approach, which relies solely on sociolinguistic cues and ignores interplay between players across multiple threads. Additionally, we restrict each player to their first \(n\) utterances and perform the same exercise. Fig.~\ref{fig:ablation} shows that the graph-aware approach achieves modest accuracy with the very first half dozen chat messages.
% \vspace{-0.15cm}

\begin{table}[!h]

% \vspace{-0.3cm}

\begin{center}
% \fontsize{9}{10}\selectfont
\begin{tabular}{l r}
\hline
\textbf{Approach} & \textbf{Accuracy}\\
\hline
SBIRL & 0.71\\
Graph-aware SBIRL & 0.79\\
\hline
\end{tabular}
% \vspace{-0.15cm}
\caption{Reward function accuracy, defined as the fraction of times the winner had a greater average estimated reward}
\label{tab:sbirl}
\end{center}

\end{table}
% \vspace*{-\baselineskip}
% \vspace{-0.85cm}
% \section{References}
% \vspace{-0.55cm}
\section{Conclusion and Future Work}
The SBIRL approach shows promising results; however, the environment lacks counterfactuals. Judging any one action is difficult because the player never chose the alternate action at that specific point in time, thus making agent-focussed testing difficult. In future experiments, we will explore:
% \vspace{-0.1cm}
\begin{itemize}
% \setlength\itemsep{0.0001cm}
% \item Further experimentation with complex, non-linear parameterizations of the reward function estimator.
\item Complex, non-linear reward function estimators.
% \vspace{-0.1cm}
\item Bootstrapping the dataset to generate counterfactual data, allowing us to build and test well-defined agents.
% \item Testing our approach on other multiparty setups.
\end{itemize}

% \vspace{-0.2cm}

% \bibliographystyle{aaai}

\bibliography{aaai22.bib}



\section*{Supplementary material}

# Supplementary materials to Using graph-aware reinforcement learning to identify winning strategies in Diplomacy games

Hansin Ahuja,[1] Lynnette Hui Xian Ng,[2] Kokil Jaidka[3]

[1]Indian Institute of Technology Ropar
[2]Carnegie Mellon University
[3]National University of Singapore
hansinahuja@gmail.com, huixiann@andrew.cmu.edu, jaidka@nus.edu.sg

## Dataset statistics

### Datasets

Table 1 provides more examples of the labeled observations in the CLAff-Diplomacy dataset. The class distributions in the data are provided in Figure 1.

Table 1: CLAff-Diplomacy labels.

| Label | Elaboration | Example Messages |
|---|---|---|
| Friendship | Exchanges or information that propel friendship between two players | "Just letting a friend know so that you can bounce him." "I'm willing to help you switch your focus to AH and Italy, and even help you eventually but we need to build the trust before I decommission the fleet in Black Sea I think." |
| Reasoning | Justify a move, guess what moves might happen next, or discuss a move that already happened | "If you took Marseilles, I would be stronger against England" "France is a really good player, and he is no doubt working hard to get England to turn on you." |
| Game Move | Actual or suggested game move | "I am committed to supporting Munich holding" "Make sure you don't move Munich so that it can take my support." |
| Share Information | Share information related to other players | "France is a really good player, and he is no doubt working hard to get England to turn on you." "And anything that's bad for Russia right now is good for Austria." |

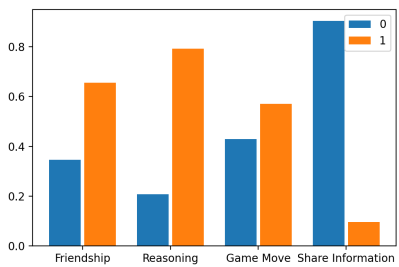

Figure 1: Dataset Composition

As noted in the main abstract, high-quality labelled data is available for only about 60% of the dataset. Further information about the entire dataset is as follows:

- Number of games: 12
- Number of chat threads: 241
- Number of utterances: 15,783

## Approach

### Weakly supervised labeling of player strategies

To predict the rhetorical strategies for the entire dataset, we relied on the following features:

- **Linguistic Inquiry and Word Count:** LIWC is a text analysis program that measures the semantic and psychological meaning conveyed in text based on a taxonomy of dictionaries (Pennebaker, Booth, and Francis 2007). LIWC has shown robust performance and captures the semantic nuances of speech, such as emotionality, attention, focus, differences in thought, and social associations and relationships.

- **Politeness strategies:** Measures of the politeness and impoliteness in speech based on the presence of lexical and stylistic features (Danescu-Niculescu-Mizil et al. 2013) as generated using the Convokit framework. The features comprise fundamental integrants of politeness theory such as deference, factuality, modality and indirection (Brown, Levinson, and Levinson 1987).

### Score-based inverse reinforcement learning

Our results from the linguistic analysis and label generation suggested distinct conversational strategies that a player might employ in a game of Diplomacy. Being able to discretize these strategies allowed us to adapt a reinforcement learning approach (by considering these strategies as actions for our RL formulation), as shown in Figure 2.

We can use inverse reinforcement learning to learn a reward function given agent demonstrations, i.e., the conversation threads between players. However, traditional formulations of inverse reinforcement learning involve demonstrations by an expert agent (Arora and Doshi 2021). One way to cast our problem into such a formulation would be to consider the winners of each game as expert agents and use their

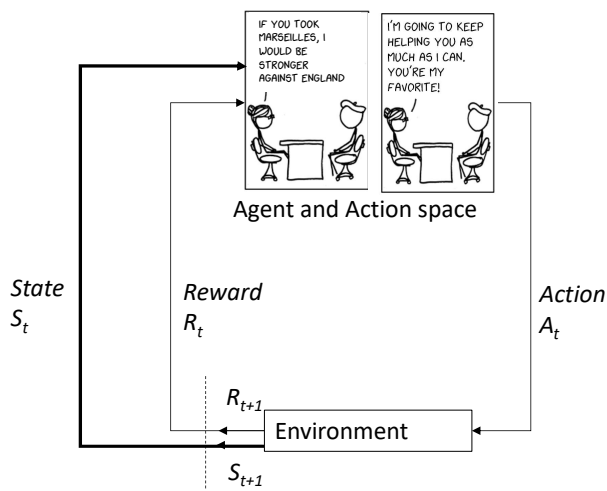

Figure 2: Reinforcement learning model for player strategies in Diplomacy.

conversations to estimate the reward function. However, this approach would lead to the elimination of almost half the dataset even if we consider the top 3 winners of each game.

Score-based inverse reinforcement learning (El Asri et al. 2016) is devised in a manner that allows us to exploit non-expert agents as well. An expert is still in play, but 1) the expert is dedicated to scoring trajectories or policies of players of all skill levels, and 2) the formulation also allows for the expert scorer to be error-prone. For Diplomacy, we have the player scores after each conversation, reflecting the success of in-game objectives such as capturing enemy territories. This score at the end of each conversation would reflect a player's success in forming successful alliances and effectively deceiving other players.

## Additional results

### Classifier performances

The full range of models evaluated for the classification tasks are reported in Table 2. The best classifiers for each label as adjudged by their accuracy on held-out data were applied to label the entire Diplomacy dataset. In subsequent experiments, we plan to train and evaluate approaches based on neural networks.

Table 2: Macro-F1 scores for feature estimators

|  | Friendship | Reasoning | Game Move | Share Information |
|---|---|---|---|---|
| Logistic Regression | 0.633 | 0.508 | 0.695 | 0.541 |
| KNN | 0.587 | 0.484 | 0.566 | 0.517 |
| Gaussian NB | 0.467 | **0.533** | 0.620 | 0.315 |
| Bernoulli NB | 0.611 | 0.517 | 0.646 | 0.547 |
| Adaboost | 0.642 | 0.448 | 0.696 | **0.623** |
| Gradient Boosting | 0.641 | 0.445 | **0.698** | 0.611 |
| Decision Tree | 0.615 | 0.511 | 0.624 | 0.591 |
| Linear SVC | 0.636 | 0.516 | 0.697 | 0.563 |
| C-SVC | **0.673** | 0.526 | 0.692 | 0.570 |
| LDA | 0.641 | 0.446 | 0.686 | 0.599 |

### Error analysis

Four runs of our approach are illustrated in Figure 3. The red line refers to the actual loser in all plots, while the green line refers to the real winner. In the true positives, an early advantage to the actual winner translated into a correct prediction. However, in the false positives shown, we notice that although the correct trends start to be captured towards the end of the trajectories, these can still not offset the advantage afforded to the losing player at the beginning of the game. In subsequent experiments, we will aim to study the temporal trends of individual games and try to study whether indicators of victory follow a temporal distribution. This would also shed light on how players may leverage the trends across time to maximize their chances of victory.

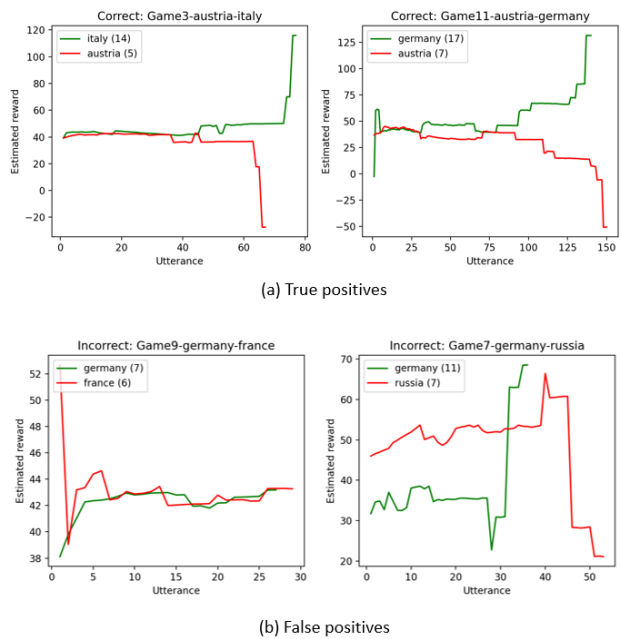

(a) True positives

(b) False positives

Figure 3: Examples of (a) True positives and (b) False positives



\end{document}